\documentclass[runningheads]{llncs}
\usepackage[table,xcdraw]{xcolor}
\usepackage{amsfonts}
\usepackage{amsmath}
\usepackage{xcolor}
\usepackage{algorithm}
\usepackage{algorithmic}
\usepackage[T1]{fontenc}
\usepackage{graphicx}
\usepackage{booktabs}
\usepackage[misc]{ifsym}

\usepackage{mwe}

\begin{document}


\title{Bridging Neural Networks and Dynamic Time Warping for Adaptive Time Series Classification}

\titlerunning{Bridging NNs and DTW for Adaptive Time Series Classification}

\author{Jintao Qu\inst{1} \and
Zichong Wang\inst{2} \and
Chenhao Wu\inst{1} \and Wenbin Zhang\inst{2} 
}

\authorrunning{J. Qu et al.}

\institute{University of Southern California, Los Angeles, CA 90089, USA \email{\{jintaoqu,wuchenha\}@usc.edu}
\and
Florida International University, Miami, FL 33199, USA \email{\{zwang114,wenbin.zhang\}@fiu.edu}
}

\maketitle              
\begin{abstract}
Neural networks have achieved remarkable success in time series classification, but their reliance on large amounts of labeled data for training limits their applicability in cold-start scenarios. Moreover, they lack interpretability, reducing transparency in decision-making. In contrast, dynamic time warping (DTW) combined with a nearest neighbor classifier is widely used for its effectiveness in limited-data settings and its inherent interpretability. However, as a non-parametric method, it is not trainable and cannot leverage large amounts of labeled data, making it less effective than neural networks in rich-resource scenarios.
In this work, we aim to develop a versatile model that adapts to cold-start conditions and becomes trainable with labeled data, while maintaining interpretability.
We propose a dynamic length-shortening algorithm that transforms time series into prototypes while preserving key structural patterns, thereby enabling the reformulation of the DTW recurrence relation into an equivalent recurrent neural network. Based on this, we construct a trainable model that mimics DTW’s alignment behavior. As a neural network, it becomes trainable when sufficient labeled data is available, while still retaining DTW’s inherent interpretability.
We apply the model to several benchmark time series classification tasks and observe that it significantly outperforms previous approaches in low-resource settings and remains competitive in rich-resource settings.

\end{abstract}

\section{Introduction}
Time Series Classification (TSC) is a subfield of machine learning concerned with assigning categorical labels to time series data. Owing to the remarkable success of deep learning in fields such as image recognition and natural language processing, researchers have increasingly explored deep learning-based approaches for TSC~\cite{NN_TSC_1,SalientSleepNet,NN_TSC_2}. Existing research primarily focuses on leveraging the parallelization and feature extraction capabilities of deep learning to improve scalability and classification accuracy. 
For instance, InceptionTime~\cite{InceptionTime}, inspired by the Inception-v4 architecture~\cite{InceptionV4}, achieves high classification accuracy while enhancing scalability through an ensemble of Convolutional Neural Networks. Similarly, Rocket~\cite{rocket} attains state-of-the-art performance by utilizing random convolutional kernels.
Despite their superior accuracy, deep learning-based TSC methods often suffer from significant limitations. In particular, they typically require large amounts of labeled data to generalize effectively~\cite{data-hungry}, which poses challenges in real-world scenarios where labeled data is scarce~\cite{pmlr-v162-zhou22f}. For example, in the medical domain, ethical constraints and data collection biases may lead to severe data imbalances, where common conditions are well-represented while rare diseases lack sufficient samples~\cite{rolando2025labeled}. Moreover, while state-of-the-art deep learning models employ complex ensembles to improve classification performance, these intricate architectures often compromise interpretability~\cite{kook2022deep}. In critical applications such as medical diagnosis~\cite{wang2022designing}, where transparency and explainability are essential, this trade-off remains a major concern.
However, deep learning is not the only effective approach. Traditional instance-based methods, such as DTW, remain highly competitive due to their robustness against temporal misalignment and ability to perform well in low-resource settings.


DTW is widely used in time series classification for its capability to align sequences with temporal distortions, for many years, the nearest neighbor classifier with DTW (NN-DTW) has been the predominant approach for time series classification\cite{bagnall2017great}. Despite extensive research efforts to develop alternative similarity measures~\cite{dtw1,dtw2,dtw3}, NN-DTW remains one of the most competitive approaches on the UCR Archive~\cite{UCR_DS}, which serves as the standard benchmark for evaluating TSC performance.
As an instance-based approach, NN-DTW provides inherent interpretability by explicitly aligning time series instances. Additionally, unlike deep learning models that require extensive labeled data, NN-DTW can be deployed with only a few instances, making it particularly effective in cold-start scenario~\cite{tapnet}. However, NN-DTW lacks trainability, which limits its ability to improve as more data becomes available. This limitation is particularly critical in domains such as medical diagnosis, where models must operate effectively in both cold-start scenarios with scarce patient data and data-rich environments where large-scale labeled datasets are progressively accumulated~\cite{meta-care}. In such settings, a adaptive approach is required—one that can ensure reliable classification under low-resource conditions while continuously improving through data-driven optimization when sufficient labeled data is available.

Incorporating the strengths of both neural networks and instance-based approaches is crucial for many applications. However, several fundamental challenges prevent existing techniques from achieving this goal, including: 
\textbf{1) Bridging Instance-Based and Neural Paradigms.} Instance-based methods such as DTW offer strong performance in low-resource scenarios, while neural networks excel in learning from large-scale data through end-to-end optimization. However, combining these advantages in a single model is non-trivial. It requires a trainable architecture that not only supports alignment-based reasoning but also enables gradient-based learning, allowing the model to adapt across diverse data regimes without sacrificing generalization or flexibility.
\textbf{2) Reducing the Computational Overhead of Alignment.} Dynamic Time Warping involves recursive dynamic programming, which is inherently sequential and difficult to parallelize. This hinders scalability and integration with modern GPU-accelerated deep learning frameworks. To bridge this gap, an effective approach must restructure DTW-like alignment operations into differentiable, parallelizable forms without losing their alignment interpretability—thus enabling fast training and inference in large-scale settings.
\textbf{3) Learning Robust Representations under Structural Variability.}
Time series often exhibit significant variability in length, local distortions, and structural patterns. While DTW handles alignment, it does not provide a robust representation that generalizes across such variations. Neural networks, on the other hand, may overfit to noise or irrelevant patterns without explicit structural guidance. Designing a representation that captures salient alignment structures while ignoring redundant or misleading variations is crucial. This requires a mechanism to extract prototypical patterns that are not only discriminative but also compact and resilient to temporal noise.




This paper proposes a universal model that preserves the structured alignment behavior of DTW while enabling trainability within a neural network framework, all while maintaining interpretability. Specifically, we introduce a length-shortening algorithm that transforms time series into prototypes while preserving key structural patterns, thereby establishing a formal equivalence between DTW and recurrent neural networks (RNNs). Building on this foundation, we design a trainable architecture that mimics DTW’s alignment process, allowing it to handle cold-start scenarios effectively while continuously improving as labeled data becomes available. At the same time, the model retains the instance-based reasoning of DTW, ensuring a transparent decision-making process and preserving interpretability.
Overall, our main contributions are as follows:
\begin{itemize}
    \item A versatile model that not only leverages DTW’s alignment properties in a trainable setting but also adapts to both cold-start scenarios with limited data and rich-resource settings, bridging the gap between instance-based methods and deep learning.
    \item A dynamic length-shortening algorithm that converts time series into prototypes while preserving key structural patterns, enabling the DTW recurrence relation to be reformulated as an equivalent recurrent neural network.
    \item A novel framework that reformulates DTW as an equivalent recurrent neural network, making it trainable while preserving its alignment behavior.
    \item Extensive experiments and case studies demonstrating the effectiveness of our approach across multiple benchmark time series classification tasks.
\end{itemize}

\section{Preliminaries} \label{Preliminaries}
Time series classification is the task of assigning a class label to a given time series based on its temporal patterns. Formally, let $\mathbf{x} = \langle x_1, x_2, \dots, x_N \rangle$ be a time series of length $N$, where each time step $x_i \in \mathbb{R}^{D} $ represents a D-dimensional feature vector. The goal of TSC is to learn a mapping function $f: \mathbb{R}^{N \times D} \to \mathcal{L}$ that assigns a label from a predefined set $\mathcal{L} = \{l_1, l_2, \dots, l_c\}$, where $|\mathcal{L}| = c$ denotes the number of distinct classes.

In distance-based methods for TSC, a fundamental challenge lies in designing effective similarity measures, as variations in sequence length, local distortions, and temporal misalignments often undermine the effectiveness of Euclidean distance. Dynamic Time Warping (DTW) mitigates this issue by computing an optimal warping path, enabling flexible alignments between time series. 

Formally, given two time series  $\mathbf{x} = \langle x_1, x_2, \dots, x_N \rangle$  and  $\mathbf{y} = \langle y_1, y_2, \dots, y_M \rangle$, DTW defines a cost matrix  $\Delta(\mathbf{x}, \mathbf{y}) = [\delta(x_i, y_j)]_{ij} \in \mathbb{R}^{N \times M}$, where each element $\delta(x_i, y_j)$  represents the alignment cost between time steps  $x_i$  and  $y_j$, typically computed using a predefined distance function, such as the squared Euclidean distance. A valid alignment between  $\mathbf{x}$  and  $\mathbf{y}$ can be represented as a warping path $p = [(e_1, f_1), (e_2, f_2), \dots, (e_s, f_s)]$ which traces a valid path through the cost matrix  $\Delta$ , connecting the upper-left entry  $(1, 1)$  to the lower-right entry  $(N, M)$  via permitted moves: downward $(\downarrow)$, rightward $(\rightarrow)$, or diagonally downward-right $(\searrow)$. 

To quantify the overall alignment cost of a given warping path, we define the DTW distance as the minimum cumulative alignment cost over all valid warping paths:
\begin{equation}\label{bg1}
dtw(\mathbf{x}, \mathbf{y}) = \min_{p \in \mathcal{A}(\mathbf{x}, \mathbf{y})} \sum_{i=1}^{s} \Delta(\mathbf{x}, \mathbf{y})[e_i, f_i],
\end{equation}
where  $\mathcal{A}(\mathbf{x}, \mathbf{y})$ denotes the set of all valid warping paths, and each alignment cost $\Delta(\mathbf{x}, \mathbf{y})[e_i, f_i]$ is accumulated along the path $p$.

Given the prohibitive complexity of exhaustively searching all possible warping paths, DTW is conventionally computed via dynamic programming. Let $h \in \mathbb{R}^{N \times M}$  denote the cumulative cost matrix, where each entry $h[i,j]$ represents the DTW distance between subsequences $x_{1..i}$ and $y_{1..j}$. The computation of $h[i,j]$  follows the recursive relation:
\begin{equation}
    h[i, j] = \Delta(\mathbf{x}, \mathbf{y})[i, j] + 
    \min \begin{pmatrix}
         h[i-1, j], \\ 
         h[i, j-1], \\
         h[i-1, j-1]
    \end{pmatrix},
\label{bg3}
\end{equation}

where $0 < i \leq N$, $0 < j \leq M$, and $h[0, :] = h[:, 0] = \infty$. The final result is $h[n, m]$. 
After computing the entire matrix, the final DTW distance is given by  $h[N, M]$, and the corresponding optimal warping path can be retrieved via backtracking.

\section{The Proposed Method}
Here we show step-by-step how we initialize a trainable recurrent neural network with sample instances and apply to time series classification.
\subsection{Optimization of DTW via Dynamic Length Shortening}
DTW-based methods for TSC often suffer from excessive computational complexity when dealing with long sequences. The traditional DTW alignment retains the full sequence length, leading to increased storage requirements and high computational costs. Furthermore, as this study seeks to reformulate DTW-based methods into an equivalent recurrent neural network, preserving the full sequence in its original form could potentially expand the parameter space, leading to a higher risk of overfitting, increased memory usage, and slower training convergence. In this context, a natural question arises: whether the sequences used as templates for DTW comparison can be shortened without compromising their discriminative capability (i.e., without significantly increasing alignment inertia).

With this in mind, we propose a dynamic length-shortening method that iteratively reduces the sequence length while preserving the essential structural characteristics required for DTW alignment. The core idea is to merge the closest successive coordinates, effectively compressing redundant information while minimizing distortions in the temporal structure, thereby achieving a more compact yet representative sequence. 



\begin{figure}
\centering
\includegraphics[scale=0.35]{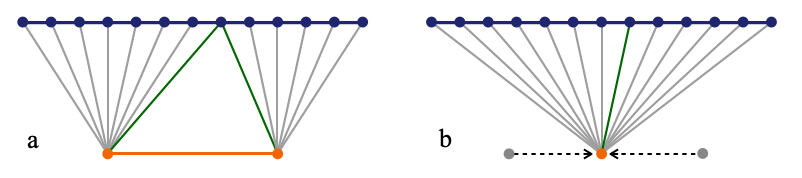}
\caption{Illustration of the dynamic length-shortening process. (a) The original alignment, where the bottom sequence contains redundant consecutive coordinates, (b) the alignment after adaptive length shortening, where redundant coordinates are merged.} \label{fig:shorten}
\end{figure}

To illustrate how dynamic length shortening works, let us consider a simple example shown in Fig. \ref{fig:shorten}. Specifically, Fig. \ref{fig:shorten}(a) illustrates the DTW alignment between two sequences, where the bottom sequence (marked in orange) initially contains two separate coordinates aligned to multiple points in the top sequence. According to dynamic length-shortening, these two coordinates can be merged into a single coordinate due to their similarity. Fig. \ref{fig:shorten}(b) demonstrates the resulting alignment after applying length shortening, where the two original coordinates in the bottom sequence are merged into one. Despite this reduction in length, the DTW alignment distance remains nearly unchanged, indicating that the shortened sequence effectively preserves the temporal information of the original sequence.

Furthermore, as illustrated by the green-marked correspondences in Fig. \ref{fig:shorten}, the shortening process tends to eliminate many instances of one-to-many alignments between the sample sequence (blue) and the prototype sequence (orange). In this case, the DTW warping path tends to consist only of downward ($\downarrow$) and diagonal ($\searrow$) moves, eliminating horizontal ($\rightarrow$) transitions. In other words, after appropriately scaling the prototype sequence, the state transition equation of DTW can be approximately optimized as Equation \ref{eq:DTW1}.

\begin{equation}
    h[i, j] = \Delta(\mathbf{x}, \mathbf{y})[i, j] + \min\begin{pmatrix}
         h[i-1, j], \\ 
         h[i-1, j-1]
    \end{pmatrix},
\label{eq:DTW1}
\end{equation}

In the context of TSC, the primary objective is to quantify the similarity between the sample sequence and the shortened prototype sequence. Consequently, it is unnecessary to retain the entire cost matrix for backtracking to determine the optimal warping path. Thus, Equation \ref{eq:DTW1} can be further reformulated into a more space efficient form, as presented in Equation \ref{eq:DTW2}.

\begin{equation}
    h_t[j] = \Delta_{t}(\mathbf{y})[j] + \min \begin{pmatrix}
 h_{t-1}[j], \\
 h_{t-1}[j-1]
\end{pmatrix},
\label{eq:DTW2}
\end{equation}
where $h_t \in \mathbb{R}^{N + 1}$ denotes the states at time step $t$, and $\Delta_{t}(\mathbf{y}) = [\delta (x_t, y_j)]_{j} \in \mathbb{R}^{L}$ denotes the pairwise distance between the sequences at step $t$. The indices satisfy $0 < t \leq N$ and $0 < j \leq L$, with boundary conditions $h_0[:] = \infty$ and $h_t[0] = \infty.$ The final cost is given by $h_N[L]$.

\subsection{From Optimized DTW to Recurrent Neural Network}
In the classification task, computing the DTW distance between a given sample sequence $\mathbf{x}$ and multiple prototype sequences from different classes requires evaluating DTW distances in a batch-wise manner. To achieve this, we define the set of prototype sequences as $S = \left \{ \mathbf{y}^{1}, \mathbf{y}^{2}, ..., \mathbf{y}^{K}\right \}$, where $K$ is the number of prototypes. The pairwise distance matrix between the sample sequence and all prototype sequences at time step $t$ is given by $\Delta_{t}(S) = [\delta(\mathbf{x}_t, \mathbf{y}_{j}^{i})]_{ij} \in \mathbb{R}^{K \times L}$. Similarly, let $h_t \in \mathbb{R}^{K \times L}$ represent the accumulated DTW cost matrix, where $ h_t[i, j]$ denotes the DTW distance between $ \mathbf{x}_{1..t}$ and $\mathbf{y}^i_{1..j}$.  

By extending Equation \eqref{eq:DTW2} to batch computation, the recursive update for DTW distances can be expressed as follows:  

\begin{equation}
      h_t[i, j] = \Delta_{t}(S)[i, j] + \min \begin{pmatrix}
      h_{t-1}[i, j], \\
      h_{t-1}[i-1, j]
      \end{pmatrix},
      \label{eq:DTW5}
\end{equation}

where $0 < i \leq K$, $0 < j \leq L$. The boundary conditions are defined as $h_t[:, 0] = \infty$, and the initialization at $t = 0$ is set as $h_0[:, :] = \infty$. The final DTW distances for classification are obtained as $h_N[:, L]$, representing the DTW costs between the sample sequence and all prototype sequences across different classes.

At this point, Equation \eqref{eq:DTW5} can be regarded as a recurrent neural network of the form $h_t = f(h_{t-1})$. In this context, the first term, $\Delta_{t}(S)[i, j]$, can be interpreted as an input-dependent transformation at time step $t$, serving as an input transformation that reflects the similarity between the sample sequence at time step $t$ and the prototype sequence features.

To integrate the DTW-based recurrence into a trainable neural network model, we represent the computation in terms of standard tensor operations. The structured recurrence defined in Equation \eqref{eq:DTW5} can be reformulated within a deep learning framework using a sequence of differentiable operators, allowing gradient-based optimization.

The proposed recurrent neural network is parameterized by a set of trainable variables, denoted as $\Theta = \langle P, h_t, O, W, b \rangle$, where: 
\begin{itemize}
    \item $P \in \mathbb{R}^{K \times L \times D}$ represents the prototype tensor constructed by directly concatenating $K$ prototype sequences. Here, $L$ corresponds to the sequence length after length shortening, and $D$ is the feature dimension at each time step. Specifically, $P[i, t] \in \mathbb{R}^{D}$ denotes the feature vector at time step $t$ of the $i-th$ prototype sequence.  
    \item $h_t \in \mathbb{R}^{K \times L}$ represents the hidden state matrix at time step $t$, corresponding to the recursive cost aggregation in Equation \eqref{eq:DTW5}. 
    \item $O \in \mathbb{R}^{L}$ is a static one-hot vector with $O[L] = 1 $, enabling the extraction of the final alignment cost. The network output is computed as $h_t O \in \mathbb{R}^{K}$, representing the final cost for each prototype.
    \item $W \in \mathbb{R}^{K \times D}$ and $b \in \mathbb{R}^{K}$ parameterize a linear transformation applied to the input signal, ensuring feature alignment before feeding into the recurrent structure.  
\end{itemize}
\begin{figure}
\centering
\includegraphics[scale=0.36]{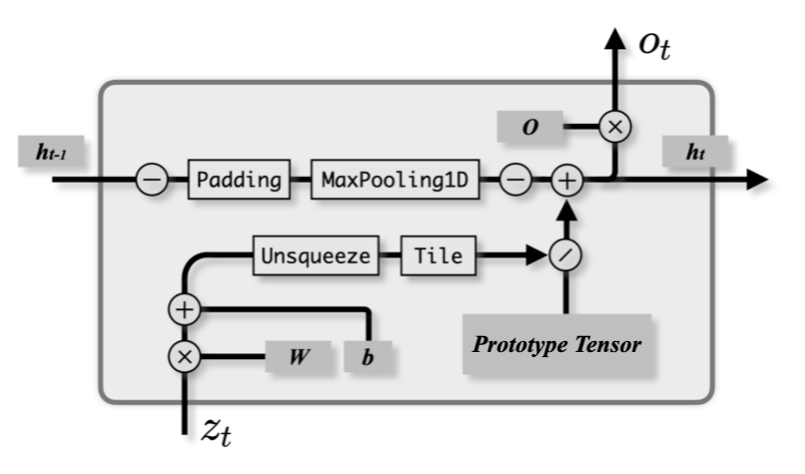}
\caption{Illustration of the computation flow in the proposed neural model.}
\label{dtw-rnn}
\end{figure}

Based on the aforementioned model definition, the overall computational flow is illustrated in Fig.\ref{dtw-rnn}. Referring to the equivalent transformation in Equation \eqref{eq:DTW5}, this computation can be decomposed into following two parts:
\subsubsection{(1) Computation of $\Delta_t(S)$.}

At time step $t$, the model takes the sample's feature representation $\mathbf{x}_t \in \mathbb{R}^{D}$ as input, along with the hidden state from the previous step $h_{t-1}$. To ensure compatibility with the prototype tensor $P$, $\mathbf{x}_t$ undergoes a linear transformation and is then broadcasted across all prototype sequences, enabling pairwise distance computation between the transformed sample representation and each prototype, which is computed as:

\begin{equation}
    \Delta_t(S) = \| (W \mathbf{x}_t + b) \mathbf{1}^{\top} - P \|^2,
\end{equation}

where \( \mathbf{1} \in \mathbb{R}^{L} \) is an all-ones vector that enables replication along the second dimension. The notation \( \|\cdot\|^2 \) represents the pairwise squared Euclidean distance computed across the feature dimension. In practice (as illustrated in Fig. \ref{dtw-rnn}), the replication induced by \( \mathbf{1}^{\top} \) can be implemented using tensor operations, where \texttt{Unsqueeze} introduces an additional dimension and \texttt{Tile} expands it along the second axis to align with the prototype tensor.
\subsubsection{(2) Computation of $\min \begin{pmatrix}
      h_{t-1}[i, j], \\
      h_{t-1}[i-1, j]
      \end{pmatrix}$.
}
At time step  $t$ , computing the minimum between  $h_{t-1}[i, j]$  and  $h_{t-1}[i-1, j]$  can be rewritten using the identity  $\min(a, b) = -\max(-a, -b)$. To ensure consistency with DTW’s alignment mechanism, we first apply a padding operation to the previous hidden state  $h_{t-1}$, producing  $h^{\prime}_{t-1} \in \mathbb{R}^{M \times (L+1)}$  with  $h^{\prime}_{t-1}[:, 0] = \infty$. This initialization follows the DTW recurrence formulation, ensuring that the boundary conditions are correctly handled.
With this transformation, the minimum operation can be reformulated as:
\begin{equation}
\begin{aligned}
    \min \begin{pmatrix} h_{t-1}[i, j], \\ h_{t-1}[i-1, j] \end{pmatrix} 
    &= -\max \begin{pmatrix} -h_{t-1}[i, j], \\ -h_{t-1}[i-1, j] \end{pmatrix} \\
    &= -\text{MaxPooling1D} \Big( -h^{\prime}_{t-1} \Big),
\end{aligned}
\end{equation}
here max-pooling is performed using a 1D max-pooling layer with a kernel size of 2 and a stride of 1, which is functionally equivalent to the max function. This design ensures that the recurrence relation of DTW is efficiently computed within deep learning frameworks while maintaining differentiability.

\subsection{Logical Aggregation Layer for Classification}
Building upon the neuralized dynamic time warping computation introduced in the previous section, we obtain a set of similarity scores between the input sequence $\mathbf{x}$ and the prototype set $S$ , where each prototype represents a reference sequence for a specific class. These similarity scores, derived from softmax-normalized DTW distances, serve as intermediate alignment measures between the input sequence and various class prototypes.

To facilitate classification, we introduce an aggregation mechanism that consolidates similarity scores into a class-level decision. Given an input sequence  $\mathbf{x}$, we first compute softmax-normalized alignment scores between  $\mathbf{x}$  and its corresponding prototypes. Each class is represented by a collection of prototypes, where  $S_{l_c} = \{y^1, y^2, …, y^k\}$ denotes the set of $k$ prototypes corresponding to class $l_c$. The classification probability is computed by aggregating these similarity scores, which can be interpreted as a differentiable approximation of logical disjunction ($\bigvee$). Prior work \cite{soft-logic} has shown that first-order logic principles can be integrated into neural networks by approximating discrete logical rules with trainable functions. Following this approach, we approximate the logical disjunction of prototype alignments using a soft aggregation function as shown in Equation \ref{eq:Score}:
\begin{equation}
Score(l_c) = \bigvee_{y^i \in S_{l_c}} \sigma\big(- h_N O \big)_{i}
 \approx \min\Big(1, \sum_{y^i \in S_{l_c}} \sigma\big(- h_N O \big)_{i} \Big),
\label{eq:Score}
\end{equation}
where $\sigma(\cdot)$ denotes the softmax function. The final predicted class is the one with the highest aggregated score:
\begin{equation}
\hat{l} = \arg\max_{l \in \mathcal{L}} Score(l).
\label{eq:final_class}
\end{equation}
This computation can be efficiently implemented using a two-layer MLP, following the framework in \cite{soft-logic}, where logical operations are approximated by neural transformations similar to those used for mapping hidden representations to label logits in traditional neural networks.

\subsection{Interpretability}
Our model inherently preserves the interpretability advantages of instance-based methods while benefiting from the flexibility of trainable representations. Unlike black-box deep learning models that rely solely on abstract feature extraction, our approach enables a direct alignment between input sequences and learned prototypes, making the decision process more transparent. Since the learned prototypes function similarly to nearest neighbors in a transformed space, classification outcomes can be traced back to specific reference sequences, allowing for clear instance-based reasoning. This makes it easier to understand why a particular sample was assigned to a given class, which is especially valuable in high-stakes applications such as medical diagnosis. Furthermore, our model allows for the visualization of learned prototypes, providing deeper insights into how it distinguishes between different categories and adapts to new data over time, which is particularly relevant in real-world applications where understanding model updates and decision boundaries is critical for reliability.

\section{Experiments}
In this section, we evaluate the proposed method through three dimensions: (1) comparative benchmarking with state-of-the-art approaches in Sect.~\ref{SOTA}, and (2) performance analysis under varying training data scales in Sect.~\ref{COLD_START}. We provided our code and complete results with standard deviations in Github, and it is available at: https://anonymous.4open.science/r/Neurlized-Dynamic-Time-Warping-CBE5
\subsection{Experimental Settings}
\subsubsection{Datasets.}To provide a broad evaluation, we assess our proposed model on all the 85 benchmark datasets from the UCR Time Series Archive\cite{UCR_DS}. Given that the effectiveness of classification algorithms varies across time series domains, we further highlight 40 datasets that are particularly relevant to distance-based methods. These datasets are characterized by temporal misalignment, dynamic pattern variability, and class separability that is better captured by similarity-based measures rather than purely statistical features. By focusing on this subset, we aim to provide deeper insights into how our method performs in scenarios where alignment-based similarity plays a critical role, allowing for a more targeted comparison against traditional baselines.
For the low-resource and full-training experiments (Section \ref{COLD_START}), we simulate cold-start conditions by randomly subsampling the training data at rates ranging from 1\% to 100\%. This enables us to examine the model’s performance under different levels of data availability. The 1\% sampling rate represents an extreme low-resource setting, where very few labeled instances are available, while the 100\% sampling rate corresponds to a fully data-rich scenario, serving as a baseline for evaluating the model’s learning capacity when ample training data is provided.
\subsubsection{Training.} All models in this study were implemented in Python 3.8 with CUDA 11.1, using PyTorch 1.10.0 as the primary deep learning framework. Additionally, some baseline models were constructed using existing time series analysis libraries - sktime and tsai, to ensure fair and reproducible comparisons. For hyperparameter selection, we adopted a grid search strategy to explore optimal configurations for most model parameters. Specifically, we varied the number of prototypes per class across \{5, 10, 15, 20\} and the sequence shortening ratio within \{0.3, 0.5, 0.8, 0.9\} to identify the optimal hyperparameters. A complete list of hyperparameter configurations and additional details are provided in the source code.
\subsubsection{Baselines.}
To ensure a comprehensive evaluation of our approach, we select a diverse range of baselines that allow us to assess its effectiveness against both classical and state-of-the-art models. Following the latest survey on time series classification\cite{TSC_Bake_Off}, we include two state-of-the-art deep learning-based methods, Rocket and InceptionTime\cite{InceptionTime}, both of which have demonstrated strong performance on diverse benchmark datasets. Additionally, since our proposed model adopts a recurrent structure, we incorporate LSTM\cite{LSTM} as a baseline to provide a direct comparison with other recurrent neural network-based approaches. Given that our model is fundamentally DTW-based, we also adopt the Nearest Neighbor classifier with DTW (NN-DTW) as a baseline, as it remains one of the most widely used methods in time series classification. To further enrich the comparison, we include two ensemble-based methods, BOSS\cite{boss} and Elastic Ensemble (EE)\cite{EE}, which have been shown to perform well on various datasets by leveraging multiple similarity measures and transformation-based features.
\subsection{Benchmarking Against State-of-the-Art Models} \label{SOTA}
\begin{figure}
\centering
\includegraphics[width=0.97\textwidth]{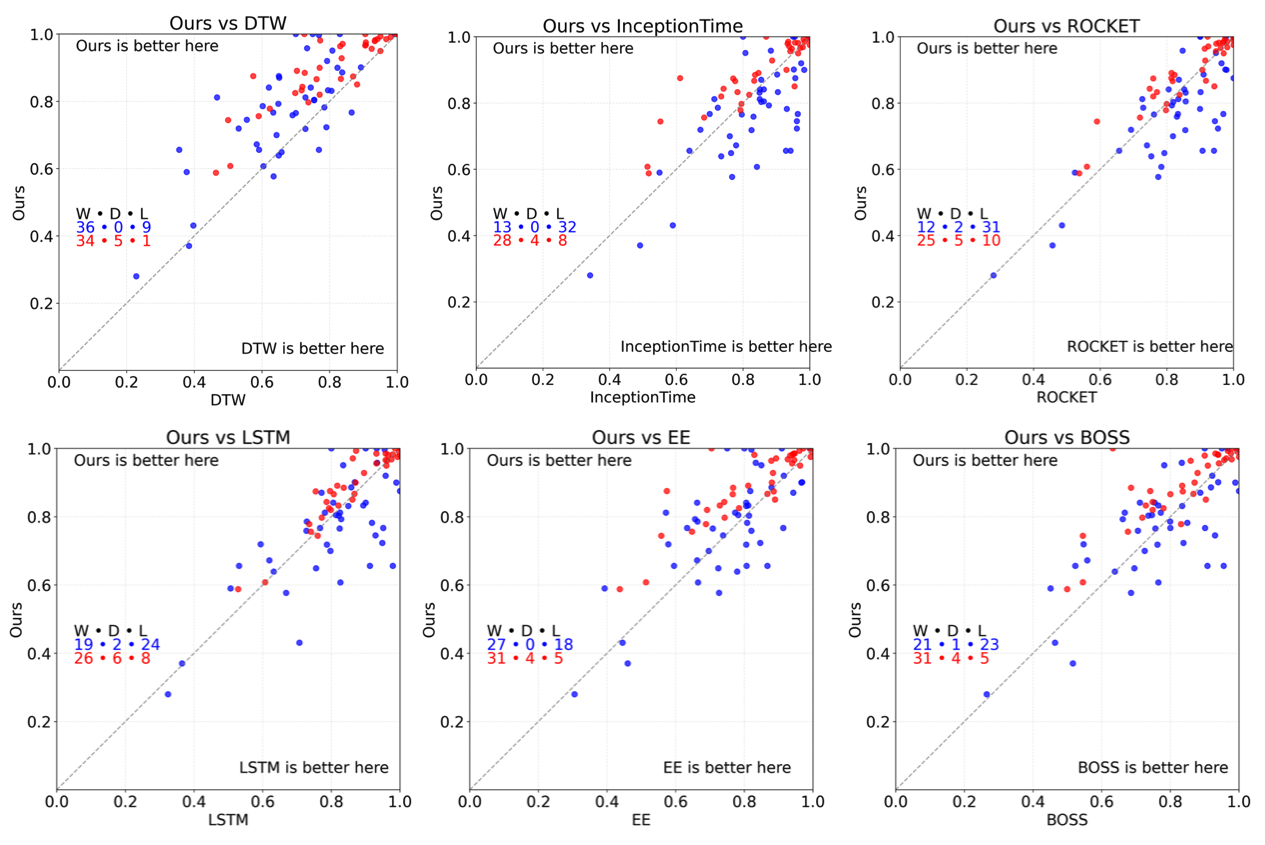}
\caption{Accuracy comparison between our model and various baselines. Each point represents a dataset, with the x-axis showing baseline accuracy and the y-axis showing our model’s accuracy. Red points indicate datasets where distance-based methods are preferred. Points above the diagonal represent cases where our model outperforms the baseline.} 
\label{fig:sota_acc}
\end{figure}
We compare our model with various baseline methods to verify its effectiveness in time series classification. For fairness, we take the best performance of each method for comparison. Fig.\ref{fig:sota_acc} presents the accuracy comparison across all 85 datasets from the UCR archive. Each point represents a dataset, where the x-axis shows the accuracy of the baseline model and the y-axis shows the accuracy of our proposed model. Points above the diagonal line indicate cases where our model outperforms the corresponding baseline, while points below the line indicate cases where the baseline achieves higher accuracy.

From Fig. \ref{fig:sota_acc}, we observe that our model demonstrates consistent superiority over DTW-kNN, particularly in distance-based datasets, winning on 34 datasets, drawing on 5, and losing on only 1, highlighting that our method effectively retains DTW’s alignment benefits while improving trainability. Against LSTM, which shares a recurrent structure, our model outperforms in 26 distance-based datasets, with 6 draws and 8 losses, demonstrating superior generalization capabilities. Comparing with deep learning-based methods, our model remains highly competitive. Against InceptionTime, we achieve 28 wins, 4 draws, and 8 losses in distance-based datasets, while against ROCKET, the numbers are 25 wins, 5 draws, and 10 losses. For ensemble-based methods, our model demonstrates strong robustness. Against EE, we achieve 31 wins, 4 draws, and 5 losses in distance-based datasets, while against BOSS, the results are 31 wins, 4 draws, and 5 losses as well, showing that our model achieves comparable accuracy with these state-of-the-art architectures. 

Overall, these results confirm that our model achieves state-of-the-art performance in distance-based datasets while remaining highly competitive across general datasets. The significant wins over DTW-kNN and LSTM validate its alignment efficiency and generalization capabilities, while the comparable results against deep learning and ensemble methods highlight its ability to balance accuracy and interpretability.
\subsection{Low-Resource and Full Training}
\label{COLD_START}
In this experiment, we simulate low-resource scenarios by randomly sampling 1\% to 100\% of the training data to examine model performance under varying degrees of data scarcity. Since the UCR archive provides predefined training splits, some datasets contain very few training samples per class, making it difficult to perform balanced subsampling across all categories. To ensure a fair and consistent evaluation, we focus on 6 datasets from UCR where each class in the training set contains at least 300 samples, allowing for controlled subsampling while maintaining class balance.

\begin{table}[]
\centering
\caption{Accuracy results in low- and rich-resource settings on six UCR datasets. Bold values indicate the best performance for each setting, while redder colors represent higher accuracy.}
\setlength{\tabcolsep}{4pt}
\scalebox{0.92}{
\begin{tabular}{@{}l*{9}{>{\centering\arraybackslash}p{0.75cm}}@{}}
\toprule
 &
  \multicolumn{3}{c}{ECG5000 (5-class)} &
  \multicolumn{3}{c}{HandOutlines (2-class)} &
  \multicolumn{3}{c}{MidPhaCorr (2-class)} \\ \midrule
\multicolumn{1}{l|}{} &
  1\% &
  10\% &
  \multicolumn{1}{c|}{100\%} &
  1\% &
  10\% &
  \multicolumn{1}{c|}{100\%} &
  1\% &
  10\% &
  100\% \\ \midrule
\multicolumn{1}{l|}{NN-DTW} &
  \cellcolor[HTML]{FFCCC9}\textbf{0.842} &
  \cellcolor[HTML]{FD6864}0.912 &
  \cellcolor[HTML]{FD6864}0.924 &
  \cellcolor[HTML]{FFCCC9}0.827 &
  \cellcolor[HTML]{FFCCC9}0.838 &
  \cellcolor[HTML]{FFCCC9}0.881 &
  \cellcolor[HTML]{B7E1CC}0.622 &
  \cellcolor[HTML]{B7E1CC}0.687 &
  \cellcolor[HTML]{DFF2E8}0.711 \\
\multicolumn{1}{l|}{InceptionTime} &
  \cellcolor[HTML]{DFF2E8}0.726 &
  \cellcolor[HTML]{FFCCC9}0.850 &
  \cellcolor[HTML]{FD6864}0.942 &
  \cellcolor[HTML]{DFF2E8}0.722 &
  \cellcolor[HTML]{FFCCC9}0.854 &
  \cellcolor[HTML]{FD6864}\textbf{0.954} &
  \cellcolor[HTML]{B7E1CC}0.694 &
  \cellcolor[HTML]{DFF2E8}0.759 &
  \cellcolor[HTML]{FFCCC9}0.818 \\
\multicolumn{1}{l|}{ROCKET} &
  \cellcolor[HTML]{DFF2E8}0.742 &
  \cellcolor[HTML]{FFCCC9}0.826 &
  \cellcolor[HTML]{FD6864}0.948 &
  \cellcolor[HTML]{DFF2E8}0.749 &
  \cellcolor[HTML]{FFCCC9}\textbf{0.876} &
  \cellcolor[HTML]{FD6864}0.943 &
  \cellcolor[HTML]{DFF2E8}\textbf{0.704} &
  \cellcolor[HTML]{DFF2E8}0.738 &
  \cellcolor[HTML]{FFCCC9}\textbf{0.842} \\
\multicolumn{1}{l|}{LSTM} &
  \cellcolor[HTML]{DFF2E8}0.702 &
  \cellcolor[HTML]{FFCCC9}0.834 &
  \cellcolor[HTML]{FD6864}0.932 &
  \cellcolor[HTML]{DFF2E8}0.700 &
  \cellcolor[HTML]{FFCCC9}0.821 &
  \cellcolor[HTML]{FFCCC9}0.886 &
  \cellcolor[HTML]{B7E1CC}0.691 &
  \cellcolor[HTML]{DFF2E8}\textbf{0.763} &
  \cellcolor[HTML]{FFCCC9}0.808 \\ \midrule
\multicolumn{1}{l|}{Ours} &
  \cellcolor[HTML]{FFCCC9}0.840 &
  \cellcolor[HTML]{FD6864}\textbf{0.916} &
  \cellcolor[HTML]{FD6864}\textbf{0.956} &
  \cellcolor[HTML]{FFCCC9}\textbf{0.832} &
  \cellcolor[HTML]{FFCCC9}0.867 &
  \cellcolor[HTML]{FFCCC9}0.892 &
  \cellcolor[HTML]{B7E1CC}0.663 &
  \cellcolor[HTML]{DFF2E8}0.756 &
  \cellcolor[HTML]{FFCCC9}0.825 \\ \midrule
 &
  \multicolumn{1}{l}{} &
  \multicolumn{1}{l}{} &
  \multicolumn{1}{l}{} &
  \multicolumn{1}{l}{} &
  \multicolumn{1}{l}{} &
  \multicolumn{1}{l}{} &
  \multicolumn{1}{l}{} &
  \multicolumn{1}{l}{} &
  \multicolumn{1}{l}{} \\ \midrule
\multicolumn{1}{l|}{} &
  \multicolumn{3}{c}{StarLightCur (3-class)} &
  \multicolumn{3}{c}{Strawberry (2-class)} &
  \multicolumn{3}{c}{Wafer (2-class)}
  \\ \midrule
\multicolumn{1}{l|}{} &
  1\% &
  10\% &
  \multicolumn{1}{c|}{100\%} &
  1\% &
  10\% &
  \multicolumn{1}{c|}{100\%} &
  1\% &
  10\% &
  100\% \\ \midrule
\multicolumn{1}{l|}{NN-DTW} &
  \cellcolor[HTML]{FFCCC9}0.841 &
  \cellcolor[HTML]{FFCCC9}0.887 &
  \cellcolor[HTML]{FD6864}0.907 &
  \cellcolor[HTML]{FFCCC9}\textbf{0.827} &
  \cellcolor[HTML]{FD6864}0.919 &
  \cellcolor[HTML]{FD6864}0.941 &
  \cellcolor[HTML]{FFCCC9}0.887 &
  \cellcolor[HTML]{FD6864}0.926 &
  \cellcolor[HTML]{FD6864}0.980 \\
\multicolumn{1}{l|}{InceptionTime} &
  \cellcolor[HTML]{FFCCC9}0.830 &
  \cellcolor[HTML]{FFCCC9}0.881 &
  \cellcolor[HTML]{FD6864}0.979 &
  \cellcolor[HTML]{DFF2E8}0.730 &
  \cellcolor[HTML]{FFCCC9}0.827 &
  \cellcolor[HTML]{FD6864}0.984 &
  \cellcolor[HTML]{DFF2E8}0.757 &
  \cellcolor[HTML]{DFF2E8}0.799 &
  \cellcolor[HTML]{FD6864}\textbf{0.999} \\
\multicolumn{1}{l|}{ROCKET} &
  \cellcolor[HTML]{FFCCC9}0.826 &
  \cellcolor[HTML]{FFCCC9}0.850 &
  \cellcolor[HTML]{FD6864}\textbf{0.981} &
  \cellcolor[HTML]{DFF2E8}0.768 &
  \cellcolor[HTML]{FFCCC9}0.854 &
  \cellcolor[HTML]{FD6864}0.981 &
  \cellcolor[HTML]{DFF2E8}0.723 &
  \cellcolor[HTML]{FFCCC9}0.811 &
  \cellcolor[HTML]{FD6864}0.998 \\
\multicolumn{1}{l|}{LSTM} &
  \cellcolor[HTML]{FFCCC9}0.822 &
  \cellcolor[HTML]{FFCCC9}0.855 &
  \cellcolor[HTML]{FD6864}0.975 &
  \cellcolor[HTML]{DFF2E8}0.754 &
  \cellcolor[HTML]{FFCCC9}0.814 &
  \cellcolor[HTML]{FD6864}0.957 &
  \cellcolor[HTML]{DFF2E8}0.707 &
  \cellcolor[HTML]{FFCCC9}0.809 &
  \cellcolor[HTML]{FD6864}0.997 \\ \midrule
\multicolumn{1}{l|}{Ours} &
  \cellcolor[HTML]{FFCCC9}\textbf{0.844} &
  \cellcolor[HTML]{FFCCC9}\textbf{0.890} &
  \cellcolor[HTML]{FD6864}0.968 &
  \cellcolor[HTML]{FFCCC9}0.816 &
  \cellcolor[HTML]{FD6864}\textbf{0.930} &
  \cellcolor[HTML]{FD6864}\textbf{0.986} &
  \cellcolor[HTML]{FFCCC9}\textbf{0.895} &
  \cellcolor[HTML]{FD6864}\textbf{0.962} &
  \cellcolor[HTML]{FD6864}0.991 \\ \bottomrule
\end{tabular}
}
\end{table}

In low-resource settings (1\% and 10\%), NN-DTW demonstrates strong performance across multiple datasets, particularly on ECG5000, Strawberry, and Wafer, where it achieves accuracy close to the highest. This is expected, as NN-DTW does not require training and can perform well even with limited labeled data. However, our model outperforms NN-DTW on four of the six datasets at the 1\% sampling rate and on all six datasets at the 10\% sampling rate, demonstrating its strong adaptability in data-scarce scenarios. Notably, our model achieves the highest accuracy in ECG5000 (10\% and 100\%), HandOutlines (1\%), StarLightCur (1\%, 10\%), Strawberry (10\%, 100\%), and Wafer (1\%, 10\%), demonstrating its robustness in different classification tasks.

In rich-resource settings (100\%), our model consistently achieves competitive or the highest accuracy compared to deep learning baselines across most datasets. Notably, in ECG5000 and Strawberry, our model significantly outperforms all baselines, demonstrating its ability to effectively utilize labeled data for training and improve prediction accuracy in data-rich scenarios. Furthermore, we observe that our model, even when trained on only 10\% of the available data, can achieve performance comparable to some fully trained baselines. This demonstrates its strong generalization ability from limited training instances, effectively bridging the gap between instance-based approaches and deep learning models. By leveraging the strengths of both paradigms, our model consistently delivers state-of-the-art performance in low-resource scenarios while remaining highly competitive in data-rich environments.

\section{Related Works}

\textbf{Low-resource Learning for TSC.} Real-world time series data is often scarce or highly imbalanced, presenting a fundamental challenge for supervised learning. Addressing this limitation is essential in domains where large-scale labeled data is difficult to obtain.~\cite{chu2024fairness}. A common approach to address this issue is data augmentation, such as over-sampling techniques~\cite{cerqueira2024time}. On the other hand, instance-based methods, such as k-NN with dynamic time warping (DTW), have been widely used for decades and naturally perform well in low-resource environments~\cite{weightedDTW}.
More recently, meta-learning techniques, such as prototypical networks~\cite{Prototypical_NW}, have been developed to mitigate the data-hungry nature of deep learning models. In the context of low-resource time series classification, prior works have explored different meta-learning strategies: cross-branch attention mechanisms for prototypical networks~\cite{RW1}, prototype embedding frameworks for capturing class discrepancies~\cite{zhang2024elastic}, and residual neural networks trained as meta-learning agents for low-resource classification. These methods generally enhance neural network performance in low-resource settings by using prototypes or extracted features to initialize model parameters.
In contrast, our approach directly reformulates DTW into a trainable neural network, leveraging its inherent suitability for low-resource scenarios while enabling adaptation through learning. This eliminates the need for prototype-based initialization and provides a unified framework that bridges instance-based and deep learning methods.

\noindent \textbf{Interpretable Neural Network for TSC.} While neural networks achieve state-of-the-art performance in time series classification, their decision-making process remains opaque. Existing interpretability methods for TSC can be broadly categorized into three types. Instance-based explanations highlight representative subsequences relevant to classification decisions~\cite{wang2019learning,early2024inherently}. Similarity-based reasoning retrieves similar time series that influenced the classification outcome~\cite{wenshedding,labaien2020contrastive}. Feature attribution methods, such as twin systems, map learned feature representations to k-NN retrieval steps, enhancing both accuracy and interpretability~\cite{Twin_1,Twin_2,Twin_3}. A comprehensive survey on these approaches is provided in~\cite{Twin_0}.
However, existing methods primarily focus on post-hoc interpretability, explaining the model’s classification behavior without allowing direct modification of the decision process. Moreover, twin-system-based approaches require labeled data and do not support manual instance specification to refine the model.
In contrast, our model provides a higher level of interpretability by not only enabling parameter visualization but also allowing fine-grained human intervention before and after training. Through our proposed model editing technique, users can modify model parameters to explicitly adjust classification behavior, making the model adaptable to evolving task specifications with minimal retraining.



\section{Conclusions}
Despite the growing interest in time series classification, existing methods often lack interpretability and struggle to generalize across both low-resource and data-rich scenarios. In this work, we address these challenges by introducing a novel interpretable recurrent neural network derived from the formal equivalence between DTW and RNN. Our approach enables instance-efficient initialization, making it effective in cold-start settings while also allowing training with labeled data to enhance predictive accuracy without compromising interpretability. To the best of our knowledge, this is the first work to neuralize DTW for time series classification. To validate our approach, we conduct extensive empirical evaluations across cold-start and data-rich settings. Experimental results demonstrate that our model outperforms existing methods in low-resource scenarios while remaining highly competitive in data-rich environments.


\begin{thebibliography}{10}
\providecommand{\url}[1]{\texttt{#1}}
\providecommand{\urlprefix}{URL }
\providecommand{\doi}[1]{https://doi.org/#1}

\bibitem{bagnall2017great}
Bagnall, A., Lines, J., Bostrom, A., Large, J., Keogh, E.: The great time series classification bake off: a review and experimental evaluation of recent algorithmic advances. Data mining and knowledge discovery  \textbf{31},  606--660 (2017)

\bibitem{cerqueira2024time}
Cerqueira, V., Moniz, N., In{\'a}cio, R., Soares, C.: Time series data augmentation as an imbalanced learning problem. In: EPIA Conference on Artificial Intelligence. pp. 335--346. Springer (2024)

\bibitem{chu2024fairness}
Chu, Z., Wang, Z., Zhang, W.: Fairness in large language models: A taxonomic survey. ACM SIGKDD explorations newsletter  \textbf{26}(1),  34--48 (2024)

\bibitem{UCR_DS}
Dau, H.A., Bagnall, A., Kamgar, K., Yeh, C.C.M., Zhu, Y., Gharghabi, S., Ratanamahatana, C.A., Keogh, E.: The ucr time series archive (2018). \doi{10.48550/ARXIV.1810.07758}, \url{https://arxiv.org/abs/1810.07758}

\bibitem{rocket}
Dempster, A., Petitjean, F., Webb, G.I.: Rocket: exceptionally fast and accurate time series classification using random convolutional kernels. Data Mining and Knowledge Discovery  \textbf{34}(5),  1454--1495 (2020)

\bibitem{early2024inherently}
Early, J., Cheung, G., Cutajar, K., Xie, H., Kandola, J., Twomey, N.: Inherently interpretable time series classification via multiple instance learning. In: The Twelfth International Conference on Learning Representations (2024)

\bibitem{LSTM}
Hochreiter, S., Schmidhuber, J.: Long short-term memory. Neural computation  \textbf{9}(8),  1735--1780 (1997)

\bibitem{NN_TSC_2}
Ismail~Fawaz, H., Lucas, B., Forestier, G., Pelletier, C., Schmidt, D.F., Weber, J., Webb, G.I., Idoumghar, L., Muller, P.A., Petitjean, F.: Inceptiontime: Finding alexnet for time series classification. Data Mining and Knowledge Discovery  \textbf{34}(6),  1936--1962 (2020)

\bibitem{InceptionTime}
Ismail~Fawaz, H., Lucas, B., Forestier, G., Pelletier, C., Schmidt, D.F., Weber, J., Webb, G.I., Idoumghar, L., Muller, P.A., Petitjean, F.: Inceptiontime: Finding alexnet for time series classification. Data Mining and Knowledge Discovery  \textbf{34}(6),  1936--1962 (2020)

\bibitem{weightedDTW}
Jeong, Y.S., Jeong, M.K., Omitaomu, O.A.: Weighted dynamic time warping for time series classification. Pattern recognition  \textbf{44}(9),  2231--2240 (2011)

\bibitem{SalientSleepNet}
Jia, Z., Lin, Y., Wang, J., Wang, X., Xie, P., Zhang, Y.: Salientsleepnet: Multimodal salient wave detection network for sleep staging. In: Zhou, Z.H. (ed.) Proceedings of the Thirtieth International Joint Conference on Artificial Intelligence, {IJCAI-21}. pp. 2614--2620. International Joint Conferences on Artificial Intelligence Organization (8 2021). \doi{10.24963/ijcai.2021/360}, \url{https://doi.org/10.24963/ijcai.2021/360}, main Track

\bibitem{Twin_0}
Keane, M.T., Kenny, E.M.: How case-based reasoning explains neural networks: A theoretical analysis of xai using post-hoc explanation-by-example from a survey of ann-cbr twin-systems. In: International Conference on Case-Based Reasoning. pp. 155--171. Springer (2019)

\bibitem{Twin_2}
Kenny, E.M., Keane, M.T.: Twin-systems to explain artificial neural networks using case-based reasoning: Comparative tests of feature-weighting methods in ann-cbr twins for xai. In: Twenty-Eighth International Joint Conferences on Artifical Intelligence (IJCAI), Macao, 10-16 August 2019. pp. 2708--2715 (2019)

\bibitem{kook2022deep}
Kook, L., G{\"o}tschi, A., Baumann, P.F., Hothorn, T., Sick, B.: Deep interpretable ensembles. arXiv preprint arXiv:2205.12729  (2022)

\bibitem{labaien2020contrastive}
Labaien, J., Zugasti, E., Carlos, X.D.: Contrastive explanations for a deep learning model on time-series data. In: International Conference on Big Data Analytics and Knowledge Discovery. pp. 235--244. Springer (2020)

\bibitem{Twin_1}
Leonardi, G., Montani, S., Striani, M.: Deep feature extraction for representing and classifying time series cases: towards an interpretable approach in haemodialysis. In: The Thirty-Third International Flairs Conference (2020)

\bibitem{soft-logic}
Li, T., Srikumar, V.: Augmenting neural networks with first-order logic. In: Proceedings of the 57th Annual Meeting of the Association for Computational Linguistics. pp. 292--302. Association for Computational Linguistics, Florence, Italy (Jul 2019). \doi{10.18653/v1/P19-1028}, \url{https://aclanthology.org/P19-1028}

\bibitem{EE}
Lines, J., Bagnall, A.: Time series classification with ensembles of elastic distance measures. Data Mining and Knowledge Discovery  \textbf{29},  565--592 (2015)

\bibitem{data-hungry}
Marcus, G.: Deep learning: A critical appraisal. arXiv preprint arXiv:1801.00631  (2018)

\bibitem{dtw1}
Marteau, P.F.: Time warp edit distance with stiffness adjustment for time series matching. IEEE transactions on pattern analysis and machine intelligence  \textbf{31}(2),  306--318 (2008)

\bibitem{rolando2025labeled}
Rolando, M., Raggio, V., Naya, H., Spangenberg, L., Cagnina, L.: A labeled medical records corpus for the timely detection of rare diseases using machine learning approaches. Scientific Reports  \textbf{15}(1), ~6932 (2025)

\bibitem{TSC_Bake_Off}
Ruiz, A.P., Flynn, M., Large, J., Middlehurst, M., Bagnall, A.: The great multivariate time series classification bake off: a review and experimental evaluation of recent algorithmic advances. Data Mining and Knowledge Discovery  \textbf{35}(2),  401--449 (2021)

\bibitem{Twin_3}
Sani, S., Wiratunga, N., Massie, S.: Learning deep features for knn-based human activity recognition. CEUR Workshop Proceedings (2017)

\bibitem{boss}
Sch{\"a}fer, P.: The boss is concerned with time series classification in the presence of noise. Data Mining and Knowledge Discovery  \textbf{29},  1505--1530 (2015)

\bibitem{Prototypical_NW}
Snell, J., Swersky, K., Zemel, R.: Prototypical networks for few-shot learning. In: Proceedings of the 31st International Conference on Neural Information Processing Systems. p. 4080–4090. NIPS'17, Curran Associates Inc., Red Hook, NY, USA (2017)

\bibitem{dtw2}
Stefan, A., Athitsos, V., Das, G.: The move-split-merge metric for time series. IEEE transactions on Knowledge and Data Engineering  \textbf{25}(6),  1425--1438 (2012)

\bibitem{RW1}
Sun, J., Takeuchi, S., Yamasaki, I.: Prototypical inception network with cross branch attention for time series classification. In: 2021 International Joint Conference on Neural Networks (IJCNN). pp.~1--7 (2021). \doi{10.1109/IJCNN52387.2021.9533440}

\bibitem{InceptionV4}
Szegedy, C., Ioffe, S., Vanhoucke, V., Alemi, A.A.: Inception-v4, inception-resnet and the impact of residual connections on learning. In: Proceedings of the Thirty-First AAAI Conference on Artificial Intelligence. p. 4278–4284. AAAI'17, AAAI Press (2017)

\bibitem{meta-care}
Tan, Y., Yang, C., Wei, X., Chen, C., Liu, W., Li, L., Zhou, J., Zheng, X.: Metacare++: Meta-learning with hierarchical subtyping for cold-start diagnosis prediction in healthcare data. pp. 449--459 (07 2022). \doi{10.1145/3477495.3532020}

\bibitem{dtw3}
Vlachos, M., Hadjieleftheriou, M., Gunopulos, D., Keogh, E.: Indexing multidimensional time-series. The VLDB Journal  \textbf{15},  1--20 (2006)

\bibitem{NN_TSC_1}
Wang, T., Liu, Z., Zhang, T., Hussain, S.F., Waqas, M., Li, Y.: Adaptive feature fusion for time series classification. Knowledge-Based Systems  \textbf{243},  108459 (2022). \doi{https://doi.org/10.1016/j.knosys.2022.108459}, \url{https://www.sciencedirect.com/science/article/pii/S0950705122001903}

\bibitem{wang2019learning}
Wang, Y., Emonet, R., Fromont, E., Malinowski, S., Menager, E., Mosser, L., Tavenard, R.: Learning interpretable shapelets for time series classification through adversarial regularization. arXiv preprint arXiv:1906.00917  (2019)

\bibitem{wang2022designing}
Wang, Y.: Designing Deep Methods to Improve Machine Learning Interpretability. Ph.D. thesis, Washington State University (2022)

\bibitem{wenshedding}
Wen, Y., Ma, T., Luss, R., Bhattacharjya, D., Fokoue, A., Julius, A.A.: Shedding light on time series classification using interpretability gated networks. In: The Thirteenth International Conference on Learning Representations

\bibitem{zhang2024elastic}
Zhang, B., Li, L., Liang, G., Tan, C., Dong, F.: Elastic slow feature prototypical network for few-shot fault diagnosis of industrial processes. IEEE Sensors Journal  (2024)

\bibitem{tapnet}
Zhang, X., Gao, Y., Lin, J., Lu, C.T.: Tapnet: Multivariate time series classification with attentional prototypical network. In: Proceedings of the AAAI Conference on Artificial Intelligence. vol.~34, pp. 6845--6852 (2020)

\bibitem{pmlr-v162-zhou22f}
Zhou, X., Liu, X., Zhai, D., Jiang, J., Gao, X., Ji, X.: Prototype-anchored learning for learning with imperfect annotations. In: Chaudhuri, K., Jegelka, S., Song, L., Szepesvari, C., Niu, G., Sabato, S. (eds.) Proceedings of the 39th International Conference on Machine Learning. Proceedings of Machine Learning Research, vol.~162, pp. 27245--27267. PMLR (17--23 Jul 2022), \url{https://proceedings.mlr.press/v162/zhou22f.html}

\end{thebibliography}
\end{document}